# Making Pose Representations More Expressive and Disentangled via Residual Vector Quantization

Sukhyun Jeong, Hong-Gi Shin and Yong-Hoon Choi
*Division of Robotics*
Kwangwoon University
Seoul, Korea
{jayze3736, ghdrl95}@gmail.com, yhchoi@kw.ac.kr

*Abstract*—Recent progress in text-to-motion has advanced both 3D human motion generation and text-based motion control. Controllable motion generation (CoMo), which enables intuitive control, typically relies on pose code representations, but discrete pose codes alone cannot capture fine-grained motion details, limiting expressiveness. To overcome this, we propose a method that augments pose code–based latent representations with continuous motion features using residual vector quantization (RVQ). This design preserves the interpretability and manipulability of pose codes while effectively capturing subtle motion characteristics such as high-frequency details. Experiments on the HumanML3D dataset show that our model reduces Fréchet inception distance (FID) from 0.041 to 0.015 and improves Top-1 R-Precision from 0.508 to 0.510. Qualitative analysis of pairwise direction similarity between pose codes further confirms the model's controllability for motion editing.

## I. INTRODUCTION

Text-to-motion for 3D human motion generation has become an essential research direction, offering broad applications in VR, animation, and robotics. By predicting human joint movements directly from text, it produces realistic motions that enhance virtual interactions, improve robotic action prediction, and streamline animation production.

Beyond motion generation itself, recent work has increasingly focused on motion editing by leveraging intermediate model representations. Among these approaches, controllable motion generation (CoMo) [1] draws inspiration from pose scripts, assigning semantic meaning to pose codes and constructing final poses by combining them. This design improves the interpretability of latent representations while enabling temporal modification of pose codes to alter poses at specific timesteps. As a result, CoMo provides both fine-grained motion control and intuitive accessibility for user-driven motion editing.

However, pose codes are inherently limited to a finite set of states, and pose latent representations are designed to represent specific keyframe poses. Consequently, a latent space formed solely by pose code combinations cannot fully capture continuous motion dynamics. To address this limitation, we propose to compute and quantize the residual between continuous latent representations and pose representations constructed from pose codes, thereby expanding the representational capacity. Moreover, since pose codes are directly used for control, their combinations must avoid unwanted interference to ensure independent and reliable manipulation. This paper introduces a learning framework designed to balance these constraints, yielding motion representations that are both interpretable and manipulable. The main contributions of this work are as follows:

- We apply residual vector quantization between pose latent representations, constructed from pose code combinations, and continuous motion features to enhance motion reconstruction performance.

- We design the model to preserve the disentanglement property of pose codes while effectively capturing fine-grained motion details.

- Through t-distributed stochastic neighbor embedding (t-SNE) visualization and pairwise direction similarity analysis, we quantitatively and qualitatively examine how disentanglement among pose codes evolves during training, providing insights into the controllability of the learned representation.

## II. RELATED WORKS

### A. Discrete Representation Learning

Vector quantized variational autoencoder (VQ-VAE) [2] has been widely applied in various domains such as image and speech synthesis [3, 4]. It proposes to quantize continuous data into discrete codes and leverage a codebook to model the data distribution in a discrete embedding space. This approach alleviates the mode collapse problem of conventional variational autoencoder (VAE) [5], providing more stable and higher-quality results.

Such methods have been extended to the motion generation domain, where several models [6, 7, 8] build upon VQ-VAE [2]. By quantizing continuous motion data into discrete representations and reconstructing them back into motion, these methods approximate the continuous motion space with a set of discrete vectors in a codebook.

T2M-GPT [6] employs VQ-VAE [2] as a motion tokenizer to represent motion as tokens, and then uses a Transformer-based model [9] to predict motion tokens from text in a two-stage procedure for 3D motion generation. This tokenization strategy has been further refined, opening opportunities to integrate language models and even large language models (LLMs).

MotionGPT [7] builds on the close relationship between motion representation and natural language. It jointly models motion tokens and text tokens, enabling bidirectional learning for both text-to-motion and motion-to-text tasks. This unified modeling allows a single framework to perform a wide range of motion-related tasks, including motion generation and description.

Parco [8] decomposes the human body into six distinct parts and models motion tokens independently for each,

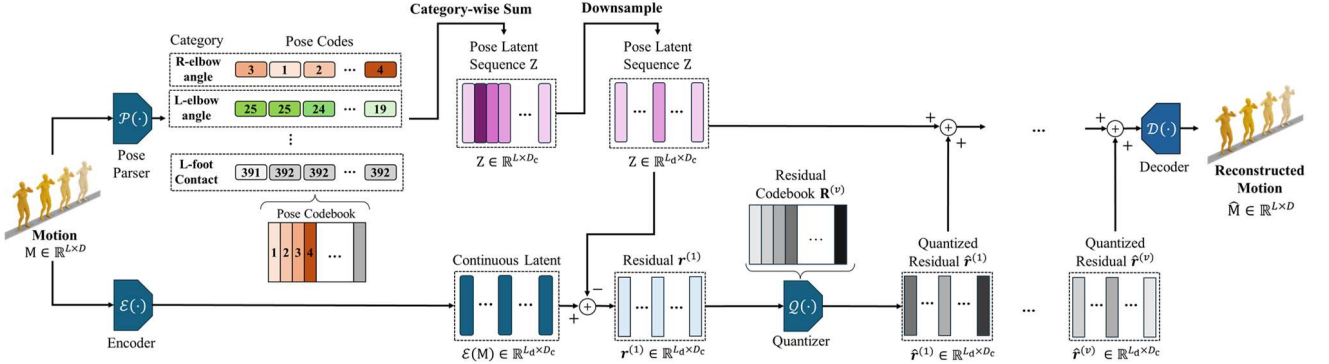

Fig. 1. Overall architecture

thereby enabling part-aware modeling that facilitates the generation of motions specific to individual body parts.

In addition to generation, recent studies have introduced models for text-based motion editing. For example, some approaches separate the upper and lower body for editing [10], while others adopt masking techniques to enable joint-level editing directly from text [11]. In these methods, users provide descriptive text about the desired modification, and a generative model such as a Transformer [9] predicts the corresponding motion tokens. The edited tokens are then used to modify the original motion accordingly.

Meanwhile, CoMo [1], inspired by PoseScript [12], proposes to assign explicit semantic meaning to pose codes that constitute the latent representation of motion. Each pose is expressed as a combination of pose codes, improving the interpretability of the latent space. This design enables intuitive motion editing, where users can manipulate specific pose codes corresponding to desired semantics. For instance, replacing the pose code "L-arm slightly bent" with "L-arm fully bent" adjusts the bending degree of the left arm. By leveraging such discrete representations learned from motion data, these approaches provide effective solutions for diverse motion-related tasks.

### B. Residual Vector Quantization

Conventional vector quantization (VQ) has long been used as an effective approach for learning discrete representations. However, the process of discretizing continuous data inevitably causes information loss [13]. Moreover, a single vector quantization step may lack sufficient capacity to effectively represent diverse motion data.

Residual vector quantization (RVQ) addresses these limitations by constructing a richer latent space capable of capturing broader contextual information without suffering from codebook collapse [14]. Compared to standard VQ, RVQ is able to represent motion features ranging from low-frequency to high-frequency components, with repeated quantization steps progressively enhancing the representation of high-frequency details [15]. The method operates by quantizing the residual error left after each quantization step and accumulating it into the existing code representation, thereby compensating for errors incurred during discretization.

Analogous to high-order spline interpolation, which approximates a continuous function using multiple spline bases, RVQ decomposes a continuous representation into multiple sets of discrete codes through iterative quantization. This process enables a more fine-grained approximation of the continuous latent space [16]. As a result, RVQ effectively converts subtle details within continuous representations into discrete codes, capturing high-frequency information and fine variations that are difficult to model with a single quantization stage.

### III. METHODOLOGY

Fig. 1 provides an overview of our framework. The input motion sequence is processed by both a pose parser and a 1D convolutional neural network (CNN) encoder to obtain the corresponding pose latent representation and continuous latent representation. The residual between them is then computed and quantized. Finally, the pose latent representation and the quantized residuals are aggregated and fed into the decoder to reconstruct the motion.

### A. Pose Codebook

Following CoMo [1], we define a pose codebook $C = \{c_n\}_{n=1}^{N} \subset \mathbb{R}^{D_c}$, where $N$ is the number of pose codes and $D_c$ denotes their dimensionality. Each pose code encodes spatial semantics, such as *L-arm slightly bent* or *L-arm fully extended*, and is mapped to one of K predefined categories. These categories capture motion-related attributes, including joint angles, inter-joint distances, relative positions and orientations, and ground-contact states.

### B. Pose Code Extraction & Pose Code Aggregation

Given a motion sequence of length $L$, $M = \{p_i\}_{i=1}^{L} \subset \mathbb{R}^{D}$, we downsample it with a stride $l$ to obtain $M_d = \{p_{i \times l}\}_{i=1}^{L_d} \subset \mathbb{R}^{D}$, where $D$ denotes the dimensionality of each pose $p_i$, and $L_d = L/l$ represents the downsampled sequence length. The downsampled motion $M_d$ is then passed into a pose parser $\mathcal{P}$, which follows the heuristic rules defined in CoMo [1] to determine whether each pose $p_i$ satisfies the semantics of a pose code $c_n \in C$. Formally, this can be expressed as: $\mathcal{P}(c_n, p_i) \to \{0, 1\}$.

As a result, the downsampled motion $M_d$ is transformed into a sequence of K-hot vectors $Z_k$, each of dimension $N$:

$$Z_k = \{\{\mathcal{P}(c_n, p_{i \times l})\}_{n=1}^{N}\}_{i=1}^{L_d}. \tag{1}$$

Next, by referencing the pose codebook, the sequence $Z_k \in \mathbb{R}^{L_d \times N}$ is converted into a pose latent sequence $Z = \{z_i\}_{i=1}^{L_d} \subset \mathbb{R}^{D_c}$:

$$Z = \{z_i\}_{i=1}^{L_d} = \left\{ \sum_{n=1}^{N} \mathcal{P}(c_n, p_{i \times l}) \cdot c_n \right\}_{i=1}^{L_d}. \tag{2}$$

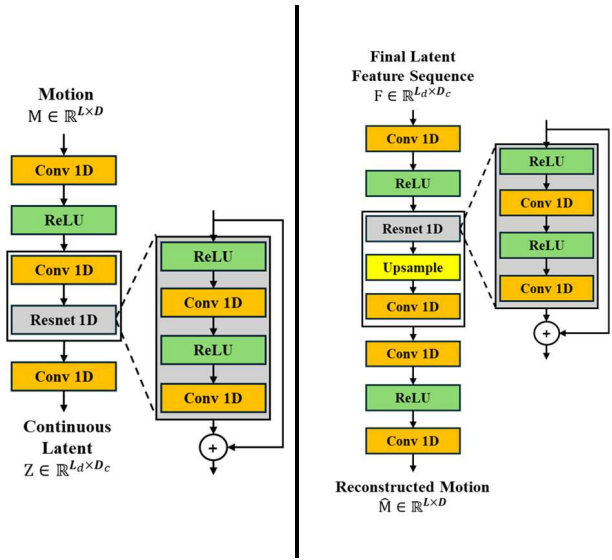

Fig. 2. Motion encoder (left) and decoder architecture (right)

## C. Residual Vector Quantization

During the construction of pose latent representations, downsampling inevitably removes certain keyframe poses, weakening temporal continuity. As a result, important pose information that explains variations at specific timesteps may be lost. This issue is particularly severe for fast actions or fine-grained movements, leading to degraded motion reconstruction performance.

To mitigate this, we apply RVQ to recover the missing details from the pose latent representations and enable richer motion modeling. A motion sequence $M \in \mathbb{R}^{L \times D}$ is first encoded by a 1D convolutional encoder $\mathcal{E}$, as illustrated in Fig. 2, producing a continuous latent representation $\mathcal{E}(M) = \{h_i\}_{i=1}^{L_d} \subset \mathbb{R}^{D_c}$. The initial residual is then computed between the encoder outputs $h_{1:L_d}$ and the pose latent representation sequence $z_{1:L_d}$:

$$r^{(0)} = h_{1:L_d} - z_{1:L_d}. \quad (3)$$

At the $v$-th quantization stage, the residual is quantized as

$$\tilde{r}^{(v)} = \mathcal{Q}(r^{(v)}) = \text{argmin}_{r_i^q \in R^{(v)}} \|r^{(v)} - r_i^q\|,$$
$$r^{(v+1)} = r^{(v)} - \tilde{r}^{(v)}, \quad (4)$$

where $\mathcal{Q}(\cdot)$ denotes the quantization process. At each stage $v = \{0, 1, \cdots, V\}$, the residual $r^{(v)}$ is mapped to the nearest code entry from the residual codebook

$$R^{(v)} = \{r_i^q\}_{i=1}^{N_r} \subset \mathbb{R}^{D_c}. \quad (5)$$

The quantized residuals $\tilde{r}^{0:V}$ are accumulated into the pose latent representations, resulting in the final latent feature sequence:

$$F = \left\{z_i + \sum_{v=0}^{V} \tilde{r}_i^{(v)}\right\}_{i=1}^{L_d}. \quad (6)$$

This process compensates for fine-grained details not captured by the original pose latent representations, yielding a more expressive motion representation. Finally, the enhanced latent features F are decoded by $\mathcal{D}$, as illustrated in Fig. 2, to reconstruct the motion $\hat{M} = \mathcal{D}(F) \in \mathbb{R}^{L \times D}$.

## D. Loss Functions

The reconstruction loss $\mathcal{L}_{recons}$ is defined as the smooth L1 distance between the input motion M and the reconstructed motion $\hat{M}$. This term enforces the model to accurately reproduce the overall structure and poses of the motion, thereby serving as the foundation for motion reconstruction:

$$\mathcal{L}_{recons} = \|M - \hat{M}\|_1. \quad (7)$$

The velocity operator $\mathcal{V}(\cdot)$ computes inter-frame velocity information from a motion sequence. Based on this, the velocity loss $\mathcal{L}_{vel}$ measures the difference in frame-to-frame variations between the input and reconstructed motions, ensuring temporal consistency:

$$\mathcal{V}(M) = \{p_{i+1} - p_i\}_{i=1}^{L-1}, \quad \mathcal{L}_{vel} = \|\mathcal{V}(M) - \mathcal{V}(\hat{M})\|_1. \quad (8)$$

The commitment loss $\mathcal{L}_{commit}$ minimizes the discrepancy between the residual $r^{(v)}$ and its quantized counterpart $\tilde{r}^{(v)}$ at each quantization stage, allowing the quantization codebooks to learn residual representations in a stable manner. In particular, during error computation at the first quantization stage, a stop-gradient is applied to the pose latent representation $z_{1:L_d}$ to avoid interfering with the alignment of pose codes:

$$\mathcal{L}_{commit} = \sum_{v=0}^{V} \|r^{(v)} - \text{sg}(\tilde{r}^{(v)})\|_2^2, \quad (9)$$

where $r^0 = h_{1:L_d} - \text{sg}(z_{1:L_d})$ and $\text{sg}(\cdot)$ denotes the stop-gradient operator.

Finally, the overall training objective is defined as a weighted sum of the three components:

$$\mathcal{L}_{final} = \mathcal{L}_{recons} + \beta \mathcal{L}_{vel} + \gamma \mathcal{L}_{commit}, \quad (10)$$

where $\beta$ and $\gamma$ are hyperparameters.

## IV. EXPERIMENT

### A. Dataset

We conducted experiments on the widely used text-to-motion dataset HumanML3D [17]. The dataset consists of 14,616 motion samples and 44,970 associated text descriptions, collected from AMASS [18] and HumanAct12 [19]. Each motion sample is paired with up to three text descriptions, and data augmentation is performed using mirroring. Motion sequences are standardized to a frame rate of 20 fps. The dataset is divided into training, validation, and test sets with splits of 80%, 5%, and 15%, respectively.

### B. Implementation Details

Our model is trained by randomly sampling a pivot point from a motion sequence and cropping a subsequence of length 64 centered at this pivot as the input. Each input sequence is temporally downsampled by a factor of 4, and two quantizer layers are used in the quantization process. The codebooks consist of two types: a pose codebook and a residual codebook. The pose codebook contains 392 vectors of dimension 512, while the residual codebook contains 64 vectors of dimension 512. In our experiments, all quantization layers share the same residual codebook.

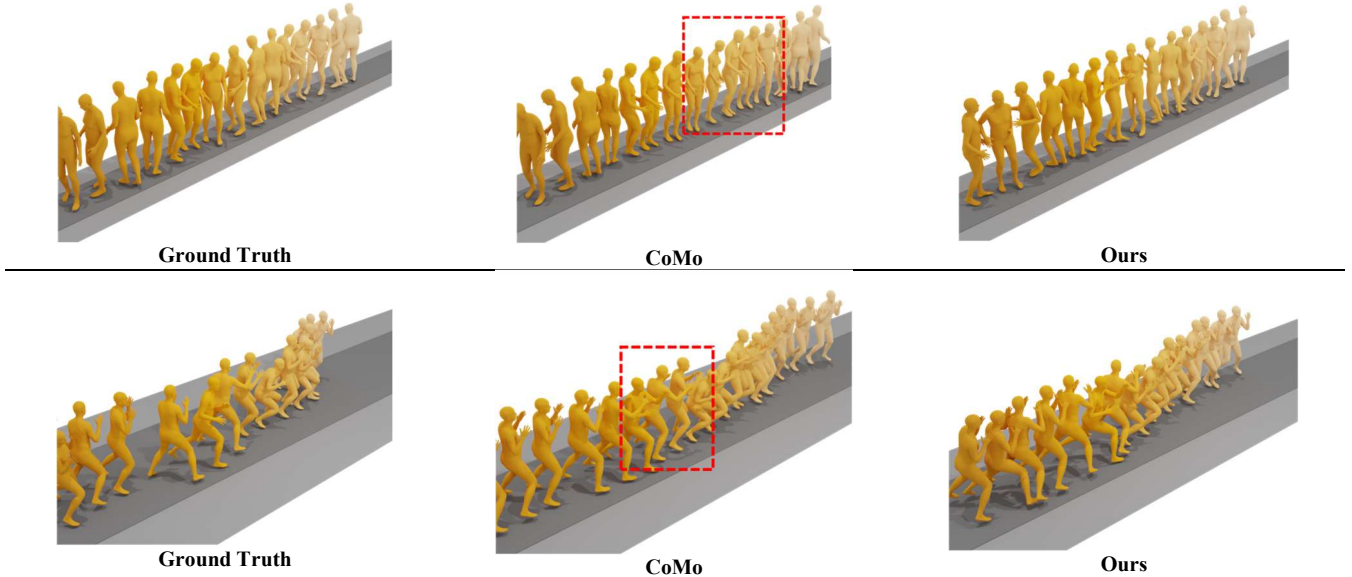

Fig. 3. Qualitative comparison with baseline models. The first row shows motion reconstruction results for the text description "a man spins in a counterclockwise circle three times with his hands in front of him." The second row shows results for the text description "a person doing specific moves with legs and hands while doing boxing."

TABLE I.   Quantitative results on the HumanML3D [3] test set. The best performance is **bold**.

| Methods | R Precision ↑ | | | FID ↓ | MM-Dist ↓ |
|---|---|---|---|---|---|
| | Top 1 | Top 2 | Top 3 | | |
| Real Motion | $0.511^{\pm.003}$ | $0.703^{\pm.003}$ | $0.797^{\pm.002}$ | $0.002^{\pm.000}$ | $2.974^{\pm.008}$ |
| T2M-GPT [6] | $0.501^{\pm.002}$ | $0.692^{\pm.002}$ | $0.785^{\pm.002}$ | $0.070^{\pm.001}$ | $3.072^{\pm.009}$ |
| Parco [8] | $0.503^{\pm.003}$ | $0.693^{\pm.003}$ | $0.790^{\pm.002}$ | $0.021^{\pm.000}$ | $3.019^{\pm.007}$ |
| CoMo [1] | $0.508^{\pm.002}$ | $0.697^{\pm.002}$ | $0.792^{\pm.002}$ | $0.041^{\pm.000}$ | $3.003^{\pm.006}$ |
| Ours | **$0.510^{\pm.002}$** | **$0.700^{\pm.002}$** | **$0.795^{\pm.002}$** | **$0.015^{\pm.000}$** | **$2.995^{\pm.006}$** |

Training is performed with a batch size of 256. The learning rate is linearly warmed up during the first 1,000 iterations, reaching a final value of $2 \times 10^{-4}$. Following MoMask [12], we adopt an exponential moving average (EMA)-based update strategy and a code reset technique to stabilize codebook training. The hyperparameters for the loss are set to $\beta = 0.5$ and $\gamma = 0.02$. All experiments are conducted on an NVIDIA A100-SXM4-80GB GPU.

### C. Metrics

We adopt the evaluation protocol proposed in T2M [17] and employ the following metrics. Fréchet inception distance (FID) measures the distributional difference between real motions and reconstructed motions, where lower values indicate better reconstruction quality. Top-R precision evaluates how well the reconstructed motion semantically aligns with its corresponding text description. Multimodal distance (MM-Dist) computes the average Euclidean distance between reconstructed motion embeddings and their corresponding text embeddings.

### D. Results

Following the evaluation protocol of T2M-GPT [6], the model selected for final testing is the one that achieves the lowest FID on the validation set. Table 1 reports the mean values and 95% confidence intervals obtained over 20 evaluation runs. Our proposed model achieves a significantly lower FID compared to the baseline, indicating superior motion representations. Furthermore, it yields a slight improvement in Top-R precision, demonstrating better semantic alignment between the generated motions and the corresponding text descriptions.

Qualitative comparisons between CoMo and our proposed model are illustrated in Fig. 3. In the first case, the text description requires the subject to perform three rotations. While the reconstructed motion from CoMo fails to complete all three rotations, our model successfully performs the full sequence. In the second case, our model produces motions with less variability than CoMo, resulting in trajectories that are closer to the ground-truth motion.

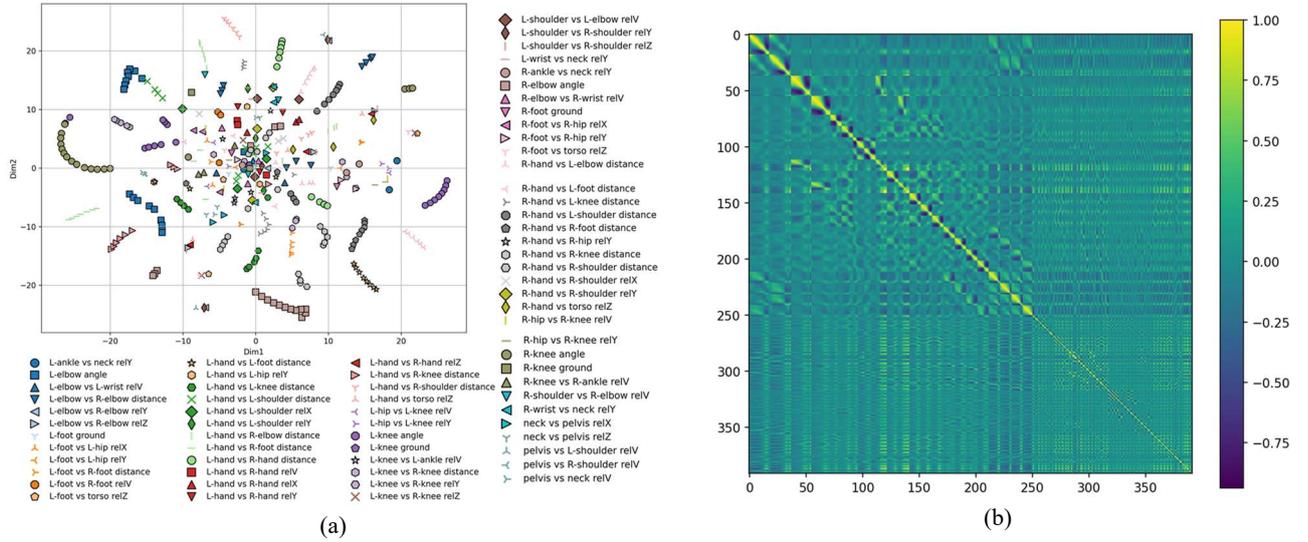

Fig. 4. (a) t-SNE projection of pose codebook entries after training. (b) Pairwise cosine similarity of pose codes after unit vector normalization.

Beyond these results, we further analyze the disentanglement property of pose codes, which is crucial for controllability. Fig. 4(a) presents a t-SNE visualization of the learned pose code embeddings, projected into two dimensions. The results show that certain categories, such as R-elbow angle and L-elbow angle, exhibit alignment patterns consistent with the regression task objective. Fig. 4(b) visualizes pairwise direction similarity by normalizing pose codes into unit vectors and computing their inner products. Pose codes within the same category tend to form subspaces in the embedding space, while those from different categories exhibit similarities close to zero. This suggests that, during motion reconstruction, pose codes implicitly maintain independence across categories.

## V. Conclusion

In this paper, we proposed a residual vector quantization framework to overcome the limited expressiveness of pose latent representations. By augmenting discrete pose codes with quantized residual features, our model preserves the interpretability and controllability of pose codes while capturing fine-grained motion details that are difficult to model with discrete codes alone. Experiments on HumanML3D demonstrated that our approach significantly improves reconstruction performance, achieving lower FID scores and higher R-Precision compared to prior methods. Furthermore, visualization and similarity analysis suggested that, while some cases reveal incomplete preservation, the disentanglement of pose codes is largely maintained, enabling controllable motion editing.

Beyond improving motion reconstruction, the proposed framework highlights the potential of combining discrete interpretable representations with residual quantization for generative modeling. We believe that this approach can serve as a strong generative prior, and future work will explore integrating our method into motion generators and extending it to broader tasks in text-to-motion, motion editing, and robotics applications.

## Acknowledgment

This work was supported in part by the Korea Agency for Infrastructure Technology Advancement under Grant RS-2025-02532980 funded by the Ministry of Land, Infrastructure and Transport under the Smart Building R&D Program, in part by the Korea Institute for Advancement of Technology under Grant RS-2024-00406796 funded by the Ministry of Trade, Industry and Energy under the HRD Program for Industrial Innovation.

## References

[1] Y. Huang, W. Wan, Y. Yang, C. Callison-Burch, M. Yatskar, and L. Liu, "Como: Controllable motion generation through language guided pose code editing," in *European Conference on Computer Vision*, 2024: Springer, pp. 180-196.

[2] A. Van Den Oord and O. Vinyals, "Neural discrete representation learning," *Advances in neural information processing systems,* vol. 30, 2017.

[3] D. Lee, C. Kim, S. Kim, M. Cho, and W.-S. Han, "Autoregressive image generation using residual quantization," in *Proceedings of the IEEE/CVF conference on computer vision and pattern recognition*, 2022, pp. 11523-11532.

[4] A. Défossez, J. Copet, G. Synnaeve, and Y. Adi, "High Fidelity Neural Audio Compression," *Transactions on Machine Learning Research*.

[5] D. P. Kingma, "Auto-encoding variational bayes," *arXiv preprint arXiv:1312.6114,* 2013.

[6] J. Zhang et al., "Generating human motion from textual descriptions with discrete representations," in *Proceedings of the IEEE/CVF conference on computer vision and pattern recognition*, 2023, pp. 14730-14740.

[7] B. Jiang, X. Chen, W. Liu, J. Yu, G. Yu, and T. Chen, "Motiongpt: Human motion as a foreign language," *Advances in Neural Information Processing Systems*, vol. 36, pp. 20067-20079, 2023.

[8] Q. Zou et al., "Parco: Part-coordinating text-to-motion synthesis," in *European Conference on Computer Vision*, 2024: Springer, pp. 126-143.

[9] A. Vaswani, "Attention is all you need," *Advances in Neural Information Processing Systems*, 2017.

[10] E. Pinyoanuntapong, P. Wang, M. Lee, and C. Chen, "Mmm: Generative masked motion model," in *Proceedings of the IEEE/CVF Conference on Computer Vision and Pattern Recognition*, 2024, pp. 1546-1555.

[11] W. Yuan et al., "Mogents: Motion generation based on spatial-temporal joint modeling," *Advances in Neural Information Processing Systems*, vol. 37, pp. 130739-130763, 2024.

[12] G. Delmas, P. Weinzaepfel, T. Lucas, F. Moreno-Noguer, and G. Rogez, "Posescript: 3d human poses from natural language," in *European Conference on Computer Vision*, 2022: Springer, pp. 346-362.

[13] C. Guo, Y. Mu, M. G. Javed, S. Wang, and L. Cheng, "Momask: Generative masked modeling of 3d human motions," in *Proceedings of the IEEE/CVF Conference on Computer Vision and Pattern Recognition*, 2024, pp. 1900-1910.

[14] H. Yao, Z. Song, Y. Zhou, T. Ao, B. Chen, and L. Liu, "Moconvq: Unified physics-based motion control via scalable discrete representations," *ACM Transactions on Graphics (TOG)*, vol. 43, no. 4, pp. 1-21, 2024.

[15] C. Wang, "T2m-hifigpt: generating high quality human motion from textual descriptions with residual discrete representations," *arXiv preprint arXiv:2312.10628*, 2023.

[16] Z. Liu et al., "CODA: Repurposing Continuous VAEs for Discrete Tokenization," *arXiv preprint arXiv:2503.17760*, 2025.

[17] C. Guo et al., "Generating diverse and natural 3d human motions from text," in *Proceedings of the IEEE/CVF Conference on Computer Vision and Pattern Recognition*, 2022, pp. 5152-5161.

[18] N. Mahmood, N. Ghorbani, N. F. Troje, G. Pons-Moll, and M. J. Black, "AMASS: Archive of motion capture as surface shapes," in *Proceedings of the IEEE/CVF international conference on computer vision*, 2019, pp. 5442-5451.

[19] C. Guo et al., "Action2motion: Conditioned generation of 3d human motions," in *Proceedings of the 28th ACM international conference on multimedia*, 2020, pp. 2021-2029.